\documentclass[letterpaper, 10 pt, conference]{ieeeconf}  % Comment this line out if you need a4paper

\IEEEoverridecommandlockouts
\usepackage{cite}
\usepackage{amsmath,amssymb,amsfonts}
\usepackage{algorithmic}
\usepackage{graphicx}
\usepackage{textcomp}
\usepackage{xcolor}
\usepackage{mathrsfs}
\usepackage{subfig}
\usepackage{multirow}
\usepackage{float}
\usepackage{booktabs}

\usepackage[acronyms, toc]{glossaries}             % Glossary, Abk und Formelzeichen. Opt. nonumberlist um Seitenz. im Glossar zu unterdrücken

\usepackage{lipsum}
\usepackage{ulem}

\title{\LARGE \bf Identifying and Extracting Pedestrian Behavior in Critical Traffic Situations  \\
}

\author{Martin Schachner$^{1*}$, Bernd Schneider$^{1}$, Fabian Weissenbacher$^{1}$, Nadezda Kirillova$^{2}$, Horst Possegger$^{2}$, \\ 
Horst Bischof$^{2}$, Corina Klug$^{1}$% <-this % stops a space
\thanks{
This work was partially funded by the Austrian Research Promotion Agency (FFG) project INTERACT (879648)
}
\thanks{$^{1}$Vehicle Safety Institute, Graz University of Technology, 8010 Graz, Austria,
        {*corresponding author: \tt\small schachner@tugraz.at}}%
\thanks{$^{2}$Institute of Computer Graphics and Vision, 
Graz University of Technology, 8010 Graz, Austria}%
\thanks{$^3$  https://doi.org/10.3217/xytxf-kjn62   }
}

\begin{document}

%
% glossary
%
\newacronym{ad}{AD}{autonomous driving}
\newacronym[plural=ADAS, firstplural=advanced driver assistant systems (ADAS)]{adas}{ADAS}{advanced driver assistant system}
\newacronym[plural=PVIs, firstplural=pedestrian vehicle interaction (PVIs)]{pvi}{PVI}{pedestrian-vehicle interaction}
\newacronym{ba}{BA}{Backward Avoidance}
\newacronym{fa}{FA}{Forward Avoidance}
\newacronym{roi}{ROI}{Region of Interest}

\newacronym{os}{OS}{Oblique Stepping}
\newacronym{nar}{NAR}{no avoidance reaction}
\newacronym{pet}{PET}{post encroachment time}
\newacronym{ttc}{TTC}{time to collision}
\newacronym{tta}{TTA}{time to approach}
\newacronym[plural=SSCs, firstplural=space-sharing conflicts (SSCs)]{ssc}{SSC}{space-sharing conflict}
\newacronym{or}{OR}{odds ratio}
\newacronym{ca}{CA}{conflict area}
\newacronym[plural=CSs, firstplural=conflict situations (CSs)]{cs}{CS}{conflict situations}
\newacronym[plural=ROIs, firstplural=Region of Interests (ROIs)]{roi}{ROI}{region of interest}

\newacronym[plural=VRUs, firstplural=Vulnerable Road Users (VRUs)]{vru}{VRU}{Vulnerable Road User}
\newacronym{aeb}{AEB}{Autonomous Emergency Braking}
\newacronym{fcw}{FWC}{Forward Collision Warning}
\newacronym{gt}{GT}{gap time}
\newacronym{pcpvi}{PC PVIs}{Potential critical PVIs}

\maketitle
\thispagestyle{empty}
\pagestyle{empty}

\begin{abstract}

% Pedestrians are reliant on partner protection. \gls{adas} are promising technologies to decrease the amount of pedestrian accidents or at least reduce collision speeds. Scenario-based evaluation is a commonly used method to assess the efficacy of new \gls{adas}. The creation of catalogs, consisting of critical pedestrian-vehicle scenarios is therefore a major challenge. One possibility to derive such a catalog are traffic observations. Besides the effort to automate the scenario reconstruction process, traffic observations benefit from a realistic determination of all relevant scenario-describing parameters, as well as the capability to detect complex and critical scenarios at a specific locations. In this paper, the inD dataset has been used as data source to extract a catalog of relevant pedestrian-vehicle scenarios. Therefore all potential pedestrian-vehicle scenarios were extracted, defined by a temporal overlap of their trajectories. The relevance of each scenario has further been assessed by evaluating the used road infrastructure as well as vehicle speed. The criticality of the remaining scenarios has been assessed, by the metrics \gls{ags}, \gls{gt}, \gls{pet}, \gls{pri}, \gls{soi}, \gls{ttz}. In order to evaluate the performance of these metrics, they have been additionally applied on vehicle-pedestrian scenarios, which are meant to be (near-)accidents by design. 

A better understanding of interactive pedestrian behavior in critical traffic situations is essential for the development of enhanced pedestrian safety systems. Real-world traffic observations play a decisive role in this, since they represent behavior in an unbiased way. 
In this work, we present an approach of how a subset of very considerable pedestrian-vehicle interactions can be derived from a camera-based observation system. For this purpose, we have examined road user trajectories automatically for establishing temporal and spatial relationships, using 110h hours of video recordings. In order to identify critical interactions, our approach combines the metric post-encroachment time with a newly introduced motion adaption metric. 
From more than 11,000 reconstructed pedestrian trajectories, 259 potential scenarios remained, using a post-encroachment time threshold of 2s. However, in 95\% of cases, no adaptation of the pedestrian behavior was observed due to avoiding criticality. Applying the proposed motion adaption metric, only 21 critical scenarios remained. Manual investigations revealed that critical pedestrian vehicle interactions were present in 
7 of those. They were further analyzed and made publicly available for developing pedestrian behavior models$^3$. 
The results indicate that critical interactions in which the pedestrian perceives and reacts to the vehicle at a relatively late stage can be extracted using the proposed method.

\end{abstract}

\section{Introduction}
\label{sec:introduction}

Every 5\textsuperscript{th} road user killed in Europe is a pedestrian~\cite{Europeancommission2021}. Following~\cite{Braess2011}, false interpretations of the environment, due to inattentiveness of the pedestrian or late reactions of the driver, e.g. due to occlusions, are the cause of 80\% of all fatalities. A recent study \cite{Casado2019} showed that infractions committed by pedestrians, such as unlawfully crossing by omitting crosswalks or walking on the road, are one of the main risk factors associated with pedestrian accidents. In other words, a pedestrian accident is the consequence of a failed \gls{pvi}.

Using the definition of \cite{Markkula2020}, \gls{pvi} can be interpreted as a traffic situation in which the behavior of the pedestrian and the vehicle is influenced by a \gls{ssc}. The term \gls{ssc} refers to the situation in which the two interacting partners would be occupying the same area at the same time in the near future. As \cite{Elvik2014} postulates, \glspl{pvi} are characterized by coordination, collaboration, competition, negotiation, and communication, which makes them relatively complex. 
In principle, there are many influencing factors affecting \glspl{pvi}, which were extensively summarized in \cite{Rasouli2020, Markkula2020}. The authors in \cite{Rasouli2020} distinguish between intrinsic (pedestrian-related) and extrinsic (environmental-related) factors.

In the definition of \cite{Markkula2020}, a proper \gls{pvi} only occurs if both the vehicle driver and the pedestrian change their trajectory. In some cases, the trajectory adaption occurs only on the pedestrian side, by the taking of a conscious action to avoid a potential \gls{ssc}, for instance by the pedestrian increasing walking speed or yielding to the vehicle. One of the most important decision criteria is the so-called \gls{gt} or \gls{tta}, on the basis of which a pedestrian decides whether or not to cross the road. The authors in \cite{Pawar2016, Gorrini2018} consider a time of 3s as the lower threshold. Of course, this always depends on the pedestrian's perception of the particular traffic situation. If this perception occurs at a very late stage, the interactive pedestrian behavior might look different. 

\begin{figure}[!ht]
  \centering
  \includegraphics[width=\columnwidth]{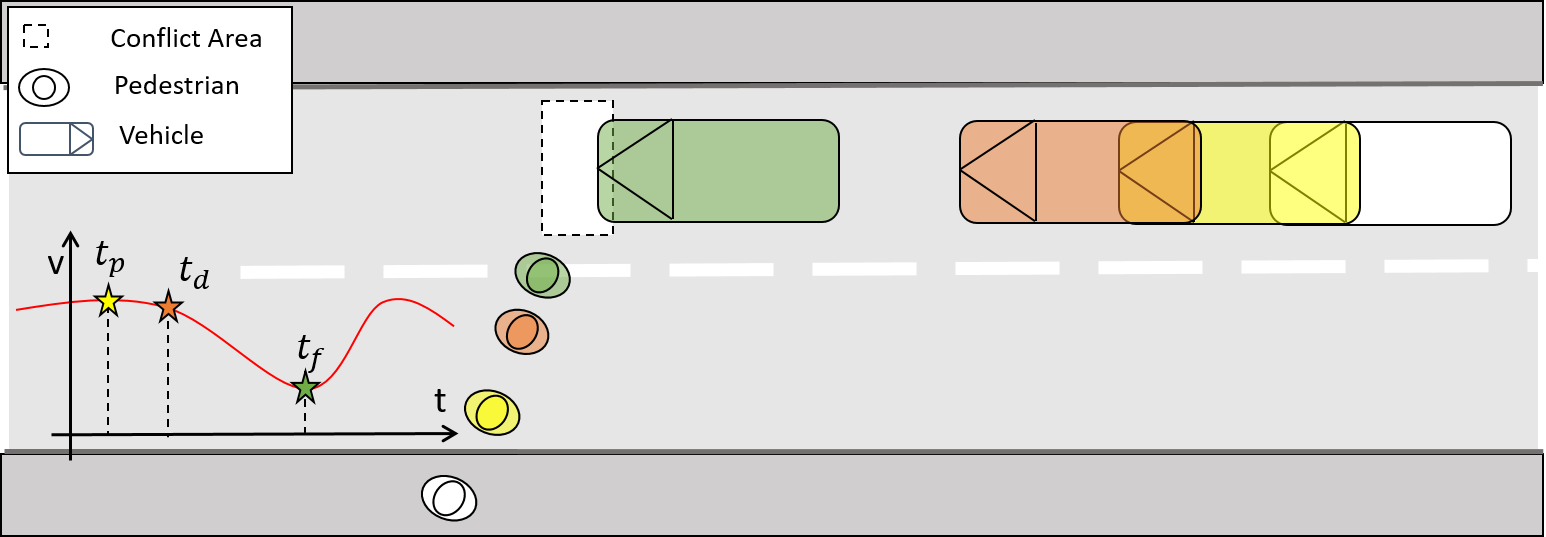}
  \caption{Example of an interactive pedestrian behavior, which is denoted by the speed profile over time. The moment in which the pedestrian perceives the vehicle is denoted as $t_p$, which is followed by a delayed reaction at moment $t_d$, where the pedestrian starts to decelerate. The $t_f$ indicates the moment when the pedestrian's deceleration is finished. From the vehicle perspective, the \gls{tta} decreases in the course of the interaction.}
  \label{fig:interactive_ped_behavior}
\end{figure}

Investigations on interactive pedestrian behavior in critical traffic situations such as those in \cite{Schachner2020a, Nie2021, Li2021} conclude that the response of the pedestrian is more of an unconscious reaction, i.e. a very abrupt change of the trajectory.  
The authors in \cite{Nie2021} conclude that upon the perception of the approaching vehicle a reaction takes place, which can be perceived as an escaping action, characterized by a changing speed profile. The studies conclude that pedestrians typically show specific reactions, such as stepping backwards or forwards, or freezing. Fig.~\ref{fig:interactive_ped_behavior} outlines the course of events for an interactive pedestrian behavior in a critical situation in which the pedestrian adapts the own speed profile at a relatively late stage by yielding to the vehicle.

% TODO: 
%   realisitischere Darstellung: 
%       Autonome Fahrzeuge aber auch für die
%       Ableiten von Unfallkonfigurationen

Enhanced pedestrian safety systems, such as \gls{adas} and \gls{ad} functions, need to correctly assess different traffic situations to possibly adapt their driving trajectory. Understanding \glspl{pvi} is therefore essential. Realistically modeling interactive pedestrian behavior that adapts to the surrounding environmentis of major improtance, especially for scenario-based development and evaluation \cite{Neurohr2020}, and in particular for the opposing traffic, as for shown for example in \cite{Yang2020}.   
% Thus it becomes clear that \glspl{pvi} are a central point for \gls{adas} to correctly assess situations and possibly adjust the driving trajectory as, for instance, shown in~\cite{Keller2014}. 
% For scenario-based development and evaluation of \gls{adas}, it is of major importance to 

Observing pedestrians in critical situations is an essential requirement for both realistically modeling pedestrian behavior and predict the outcome of \glspl{pvi}. Camera-based traffic observations are a promising method for recording multiple different \glspl{pvi} as shown e.g. in \cite{Gorrini2018}. A major advantage lies in the possibility to determine relevant influencing factors and to observe real-world scenarios, as shown for example in \cite{Kirillova2022}.

The objective of this study is to identify critical \glspl{pvi} from a pool of recorded data and extract interactive pedestrian behavior. In particular, we analyze long-term recordings from a traffic monitoring camera and propose a schema to automatically extract data sets representing critical \glspl{pvi}. These recorded \glspl{pvi} can be used to calibrate and validate pedestrian models used in traffic simulations. To this end, Section~\ref{sec:data_set} summarizes properties of the data collected, outlines preprocessing steps taken and discusses the scenario representation. In Section~\ref{sec:method}, the identification of critical \glspl{pvi} is explained. Section~\ref{sec:results} shows the obtained results in detail.

\section{DATA SET}
\label{sec:data_set}

% In recent years, many different pedestrian data sets recorded for specific applications have been published. They reach from benchmarks for tracking algorithms, \emph{e.g.}~\cite{Chavdarova2017}, to first-person vehicle centered data sets incorporating underlying road network \cite{Houston2020, Caesar2020} for \gls{adas} and \gls{ad} development to traffic observations (TODO). 
% The data sets \cite{Robicquet2016, Zhan2019, Yang2019a, Bock2020} feature \glspl{pvi} recorded by drones. A major drawback is that the viewpoints do not provide the higher level of detail  which is necessary to extensively evaluate \glspl{pvi}. The camera placement and collected data in \cite{Kirillova2022, Schachner2023} differs in this aspect and thus allows the reconstruction of further details, such as pedestrian age or gender.

\subsection{Data collection}

\begin{figure}[ht!]
  \centering
  \includegraphics[width=\columnwidth]{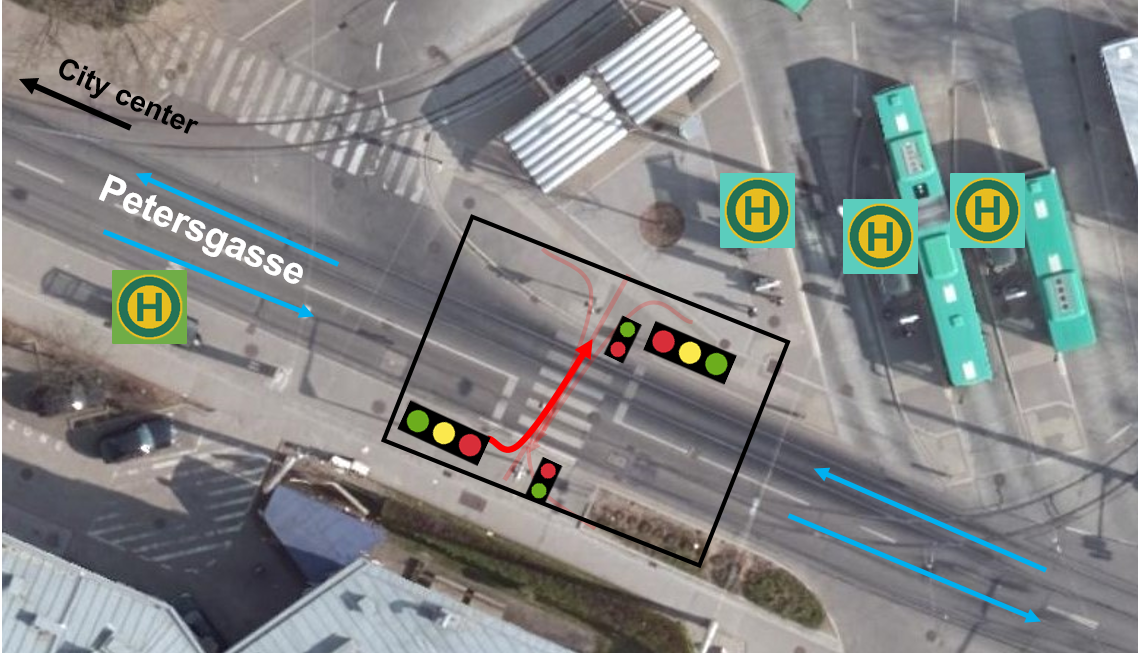}
  \caption{Orthophoto of the selected observation point St. Peter Schulzentrum. The observed traffic site is a signal-controlled pedestrian crosswalk, where pedestrians frequently rush to catch buses and trams.}
  \label{fig:observation_point}
\end{figure}

In this study, road user trajectories were collected at a signalized pedestrian crosswalk using the camera-based observation system described in \cite{Kirillova2022} and \cite{Schachner2023}. The observed crossing is located at St. Peter Schulzentrum, one of the main transfer points in Graz, Austria. Passengers changing from bus to tram (and vice versa) have to cross the main road as shown in Fig.~\ref{fig:observation_point}. Although a signal-controlled crosswalk is provided, it is often ignored by pedestrians rushing to catch public transport vehicles. The infrastructure and layout of the road are therefore likely to yield critical \glspl{pvi}, which is why it was selected as an observation point.  This assumption is further underpinned by the fact that five pedestrian accidents with personal injuries occured in the vicinity of the observed crossing within the time period 2014--2020. 
For this study, 10 days were randomly selected between May/2022 and August/2022, on which traffic was observed between 7 am and 6 pm, which resulted in 110 hours of video footage.

% \begin{figure*}[h]
%     \centering
%     \subfloat[Bird's eye view of the traffic site]{
%         \includegraphics[width=1.0\columnwidth]{img/StPeter_CS_birdseye.png}
%       }%
%     \subfloat[3D scene representation with pedestrian zones and the \gls{roi}]{
%       \centering
%         \includegraphics[width=1.0\columnwidth]{img/St_Peter_zones___.png}
%     }
%     \caption{Road layout of the traffic site of the observed pedestrian crosswalk at St. Peter Schulzentrum. (a) Aligned video frames to a bird's eye view. The highlighted arrows indicate the moving direction of the pedestrian (red) and the vehicle (black), which forms a conflict situation in which the vehicle is approaching from the near side. Other considered conflict situations are transparent. (b) Aligned 3D scene representation of the traffic site. The grey area represents the considered \gls{roi} for \glspl{pvi}. The areas in green and red highlight the approaching and target zones for pedestrians.}
%     \label{fig:observation_point}
% \end{figure*}

\subsection{Preprocessing and scenario representation}

The trajectories of pedestrians and the vehicle types under consideration (cars and bicycles) interacting with them were inferred using the method described in \cite{Kirillova2022}. The mapping to a scene representation has been done as described in \cite{Schachner2023}. A detailed model of the road network has been created for this purpose, which is shown together with reconstructed road user trajectories in Fig.~\ref{fig:observation_point_digital}. 
In order to exclude reconstructed trajectories that are not in the vicinity of the signal-controlled crosswalk, reconstructed trajectories have been examined for a spatial relationship to a defined \gls{roi}. Further, the reconstructed trajectories are time-stamped and allow the examining of a temporal relationship between pedestrians and vehicles. This spatial and temporal relationship, which was used to derive a scenario catalog of potential \gls{pvi} scenarios, is described in the following. 

\subsubsection{Spatial relationship}
\begin{figure*}[ht!]
  \centering
  \includegraphics[width=\textwidth]{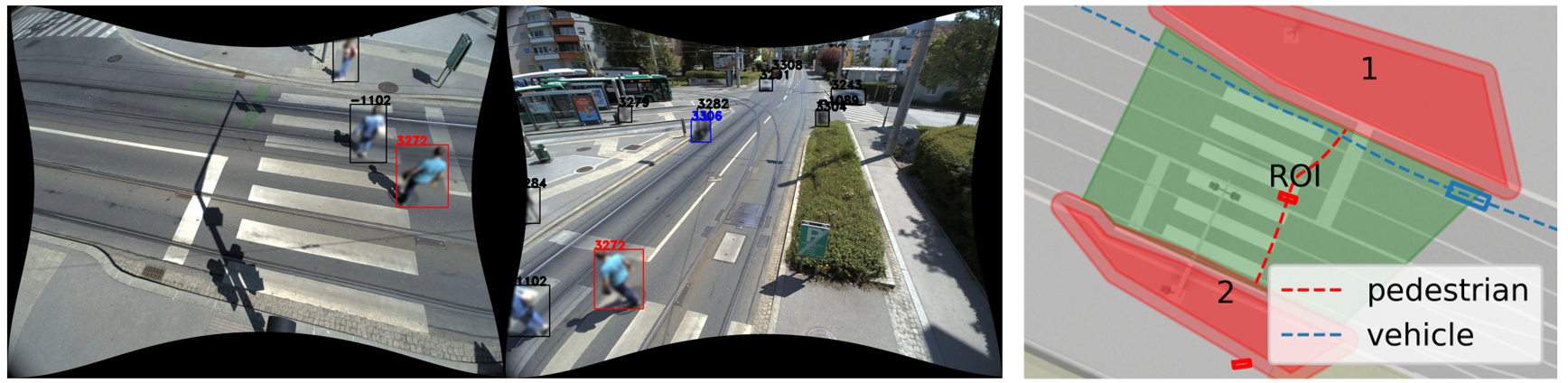}
  \caption{Aligned 3D scene representation of the traffic site. The areas in red denote the pedestrian zones (which can either be approach and target), the green area represents the \gls{roi} for \glspl{pvi}.}
  \label{fig:observation_point_digital}
\end{figure*}

Pedestrian motion at crossings is denoted by approach and target zones \cite{Gorrini2018}. For the observed traffic site, these particular zones have been defined in compliance with the road layout. By using the geometric overlap between the trajectory and the respective zones it was determined whether the pedestrian $P_i$ walked from one zone to another. In order to identify \glspl{pvi} only in the region of the crosswalk, the concept of a \gls{roi} was introduced, which is defined by the road-facing boundaries of the pedestrian zones. In a similar way, the movement of the vehicle and thus the \textit{conflict situation}, meaning whether the vehicle approached from the \textit{near-} or \textit{far-side} lane, can be determined. 
Pedestrian zones and the \gls{roi} are shown in Fig.~\ref{fig:observation_point_digital}. The reconstructed paths show a conflict situation in which the pedestrian moves from zone 2 to zone 1 and the vehicle $V_j$ approaches from the far-side lane.

\subsubsection{Temporal relationship}

For \gls{pvi} identification, pedestrians are at the center of consideration. The timeline of each scenario is described by the entrance of the pedestrian $P_i$ at time $t_s^{P_i}$ and its exit from the scene at $t_e^{P_i}$. All moving vehicles $V_j$ ($ \exists t \in [t_s^{P_i}, t_e^{P_i}] | v_j(t) \neq 0$) which have a temporal overlap with the pedestrian $P_i$ are considered as potential candidates for a \gls{pvi}, and are thus considered as the \textit{baseline catalog}.

% Scenario Property
% *****************
% Surrounding other road user? 
%   time overlapp with ego?

% Relation Ego - Obstacle 
% *********************
% relative_movement_ego_to_obstacle, ??

\section{METHOD}
\label{sec:method}

% In this section, the method of the 
% In a first step, the catalog of potential \gls{pvi} scenarios was examined for \glspl{ssc}. In a second step, pedestrian motion has been analysed for non-ordinary pedestrian motion. Subsequently, the extracted attributes were further investigated for critical relations to the conflict vehicle. 
%Der extrahierten catlaog an potentiell kritischen Scenarien, wie in xx identifiziert wurde weiterhin auf die Um auf kritsche   

% \subsection{Criticality assessment}
% \label{ss:space_sharing_conflict_and_criticality}
% --------------------------------------------------
% Welche Metrik wird verwendet, um relevante Szenarien zu extrahieren, bzw welche anderen Faktoren sind 
% Ausschlusskriterien ?
% --------------------------------------------------

% The authors in \cite{Junietz2017} and \cite{Kruber2019} have shown approaches for determining critical highway scenarios retrospective from a pool of recorded scenarios. The criticality was predominantly assessed by calculating the \gls{ttc}. In general pedestrian vehicle interactions are more complex than highway scenarios and take place commonly in urban areas with a diverse environment, consisting of traffic rules and other road user. Therefore metrics like \gls{ttc} are not well applicable and there is a need for investigation on further metrics which are suitable for pedestrian vehicle scenarios \cite{Westhofen2021}, \cite{Mahmud2017}.

% Q: For each metric, does it identify the same vehicle as most relevant for the pedestrian?  

% Literature Review based on \cite{Westhofen2021}.

The criticality assessment of a scenario depends to a great extent on the context of the application \cite{Westhofen2021}. In our approach, we combine metrics, capable of predicting a space sharing conflict with a newly proposed metric describing the pedestrian motion adaptation. With this approach we are able to identify critical \glspl{pvi} within the extracted baseline catalog.

%To search for critical \glspl{pvi} within the catalog, we used the \gls{pet} \cite{Allen1978} to determine \glspl{ssc} in a first step. 
%In principle, the \gls{pet} already provides an indicator for the proximity of the vehicle and the pedestrian. Nevertheless, there are other factors, such as the \gls{tta}, which determine the criticality of a scenario. For a pedestrian-focused consideration especially the motion adaptation within the road, as shown in Fig.~\ref{fig:interactive_ped_behavior}, represents an important characteristic. In our assessment method, this motion adaptation was further considered as a criticality metric.

%Nevertheless, there are a number of constellations which do not entail any further criticality, such as deliberately crossing after a passing vehicle has come to a standstill, or crossing just in front of it.
%A pedestrian-side consideration is for instance given by an adaption of the motion, as described in \cite{Nie2021}. Especially the flinching can be detected as an essential feature, and therefore seems to be of importance.

\subsection{Space sharing conflict}

Important metrics to quantify the criticality of \glspl{ssc} are \gls{pet} and \gls{tta} as described in \cite{Johnsson2018}. In our approach, the \gls{pet} has been used to assess the temporal proximity of pedestrians and vehicles, and the \gls{tta} has been calculated for specific \glspl{pvi} identified to be critical. %\gls{}

\subsubsection{\Acrfull{pet}}
\begin{figure}[ht!]
  \centering
  \includegraphics[width=\columnwidth]{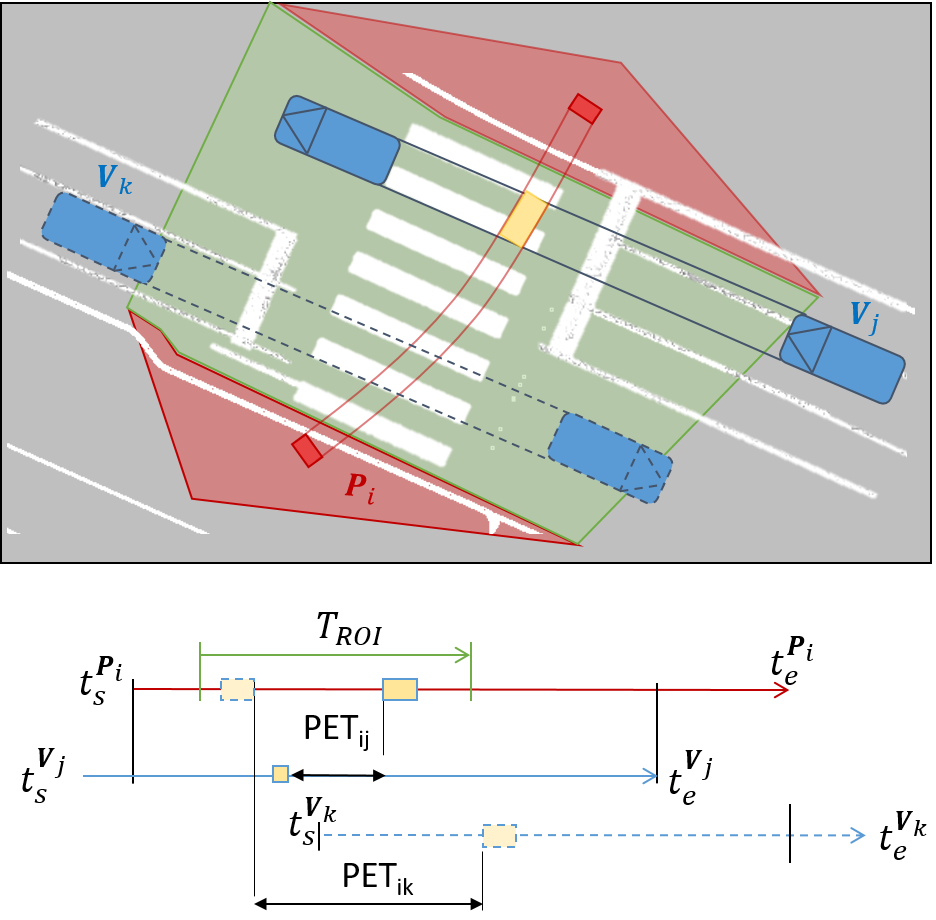}
  \caption{The red boxes represent the pedestrian $P_i$, the blue boxes the vehicle $V_j$. The yellow area, in which the two contours overlap is the \gls{ca}. For this specific conflict between $P_i$ and $V_j$, the pedestrian is moving from zone 2 to 1, and the vehicle is approaching from the far-side lane. The \gls{pet} is negative as the vehicle approaches first. For the vehicle $V_k$ the pedestrian is in the \gls{ca} before the vehicle approaching from the near-side lane.}
  \label{fig:scenario_timeline}
\end{figure}

The \gls{pet} quantifies the time gap between one road user exiting a designated \gls{ca} and another entering it. \gls{pet} has been shown to be an appropriate metric to identify \glspl{ssc} for instance by \cite{Peesapati2013, Johnsson2018}. In the given example of Fig.~\ref{fig:scenario_timeline}, the pedestrian $P_i$ passes the \gls{ca} (highlighted in yellow) after the detected vehicle $V_j$. For the vehicle $V_k$ the pedestrian is in the \gls{ca} before the vehicle. The \textit{constellation} between the $P_i$ and $V_j$ is denoted as \textit{vehicle first}, whereas the other is denoted as \textit{pedestrian first}. The formula used to calculate the \gls{pet} is outlined in the following and yields negative \glspl{pet} for \textit{vehicle first} constellations:

\begin{eqnarray}
 & \text{PET}(P_i, V_j, CA) = &  \\ 
 & \begin{cases}
  t_{entry}(V_j, CA) - t_{exit}(P_i, CA), \text{if pedestrian first} \\
 t_{exit}(V_j, CA) - t_{entry}(P_i, CA), \text{if vehicle first.} &
 \end{cases}
\end{eqnarray}

The proximity measure provided by the \gls{pet}  is an indicator of the criticality of a certain \gls{pvi}. In the further consideration only \glspl{pvi} in which the \gls{pet} had an absolute value below 2s have been taken into account.

\subsubsection{\Acrfull{tta}}

The \gls{tta} is one of the major criteria which guide pedestrians in making their cross / don't cross decision \cite{Pawar2016, Rasouli2020}. It is a predictive metric, which indicates the time when a certain vehicle would enter the \gls{ca}. Therefore, the simple relation between the vehicle's distance to the \gls{ca} $d_{V_j}(t, CA)$ at a certain moment $t$ is set in relation to the speed $v_{V_j}(t, CA)$,

\begin{equation}
%\text{TTA} =   (v_{t_i} - \hat{v}_{t_i})^2
\text{TTA}(V_j, CA, t) =  d_{V_j}(t, CA)~/~v_{V_j}(t, CA).  
\label{eq:time_to_approach}
\end{equation}

Due to the predictive nature of the metric, the \gls{ca} resulting from the retrospective view cannot be used. In our case, we consider the time $t_d$ at which the pedestrian trajectory shows a speed profile change, i.e. a deceleration, schematically represented in Fig.~\ref{fig:interactive_ped_behavior}. The pedestrian state at $t_d$ is used to linearly extrapolate its movement. From the extrapolated trajectory, an overlap of the road users is determined, which defines $\text{CA}^\prime$. For this new $\text{CA}^\prime$, the distance $d_{V_j}(t, \text{CA}^\prime)$ is calculated at each time step $t > t_d $. 
The vehicle speed $v_{V_j}(t, \text{CA}^\prime)$ results from the reconstructed trajectory of the vehicle.

\subsection{Pedestrian motion adaption}

Pedestrian motion at crosswalks has been investigated, among others by \cite{Goldhammer2014, Gorrini2018}. The findings are that pedestrian speeds are generally higher here than on dedicated sidewalks. However, these studies did not focus on critical \glspl{pvi} as done for example by \cite{Li2021, Nie2021}, who postulate that an abrupt deceleration in the \gls{roi} becomes an essential aspect in avoiding an impending collision. 

We therefore postulate that an ordinary motion profile (not triggered by an impending collision) as described in \cite{Gorrini2018}, can be approximated through a quadratic polynomial

\begin{equation}
 \hat{v}_P(t) = \beta_3 t^2 + \beta_2 t + \beta_1.
\label{eq:quadratic_function}
\end{equation}

within the period $T_{ROI}$. For the considered \glspl{pvi}, the parameters $\beta_j, j \in [1,2,3]$ are determined by curve fitting in the least square sense for each pedestrian speed profile $v_P(t)$ by optimizing 

\begin{equation}
\underset{\beta}{\operatorname{argmin}} \sum^{m}_{i = 0}  (v_{t_i} - \hat{v}_{t_i})^2.
\label{eq:curve_fitting}
\end{equation}

\begin{figure}
    \centering
    \includegraphics[width=\columnwidth]{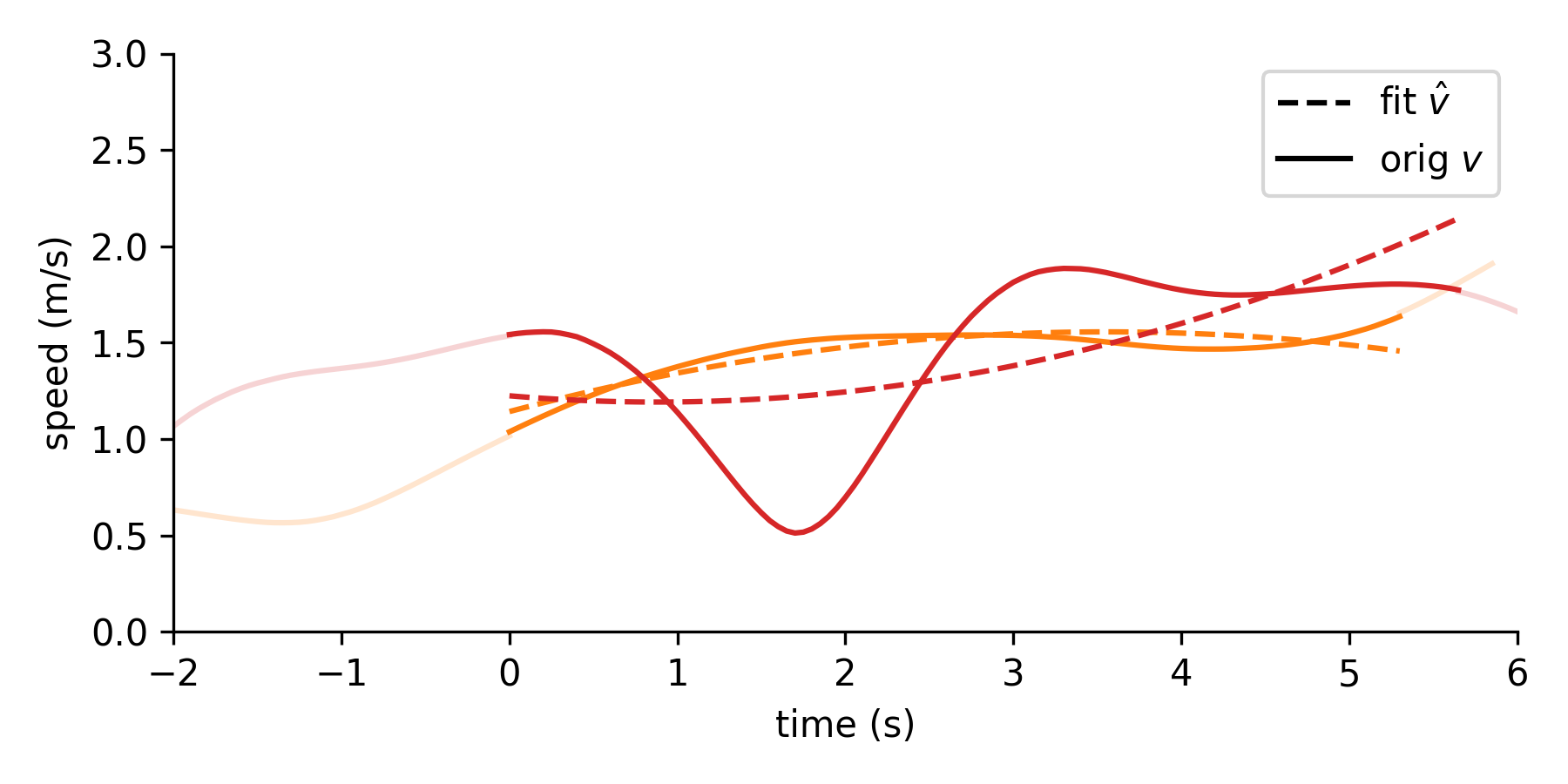}
    \caption{Exemplary curve fitting results in the least square sense, on two different pedestrian speed profiles of the reconstructed trajectories. The solid red line represents a pedestrian who is adapting the speed profile used, while the orange line presents an ordinary walking behavior. The $\operatorname{std}(\mathbf{e})$ is 0.049m/s for the non-ordinary case and 0.0067m/s for the other. The considered time period $T_{ROI}$ is indicated by the fully opaque lines, the faded parts represent the speed profile within the approach and target zone. The approximation $\hat{v}$ is outlined as dashed line.}
    \label{fig:curve_fitting}
\end{figure}

%In contrast speed profiles, induced by an abrupt deceleration would require a higher order polynomial for approximation. Therefore the fit in the least-square sense yields also a higher deviation between the quadratic approximation and the measured speed profile.
For critical \glspl{pvi}, by contrast, Equation \eqref{eq:quadratic_function} leads to a higher deviation as it is unable to represent their characteristic changes from deceleration to acceleration.
The standard deviation of the error vector $\mathbf{e} = \mathbf{v} - \hat{\mathbf{v}}$ can therefore be used as an indicator for non-ordinary pedestrian motion. To classify whether the speed profile is ordinary or non-ordinary, we have set the threshold to the 95th percentile of the $\operatorname{std}(\mathbf{e})$ overall extracted pedestrian trajectories. To outline the approach, the quadratic fit has been applied on two different speed profiles, which we identified as representative for both classes shown in Fig.~\ref{fig:curve_fitting}.

\section{RESULTS}
\label{sec:results}

In the following, we summarize properties of the extracted trajectories, described in Section~\ref{sec:data_set}. The extracted subset of \glspl{pvi}, by means of a \gls{ssc} and the proposed motion adaption metric are further part of this Section.

\subsection{Extracted trajectories}

% Pedestrian Trajectories: ~11.000     Overall, this corresponds to a distance of 78.5 (km)... from A to B and vice versa B to A ... 
% Vehicle Trajectories: ~40.000       Overall, this results in a distance of 411.9 (km) in the \gls{roi}
% Bicycle Trajectories: ~4.500        This results in a total distance of 50.9 (km) in the \gls{roi}. 

A total of 110h of recordings was analyzed in the presented study, resulting in 11,089 reconstructed pedestrian trajectories, following the requirements to intersect the pedestrian zones. 3,273 trajectories were assigned to be from zone \textit{1-2} and 7,816 from zone \textit{2-1}. The motion adaption metric was calculated for all those trajectories and resulted in a mean of 0.011($\pm$ 0.019)m/s with a 95th percentile of 0.0395m/s.
For the vehicle trajectories, 39,631  \textit{car} trajectories have been reconstructed, the distribution between \textit{out of city center} and \textit{into city center} is 21,012 to 18,619. For \textit{bicycles}, there are a total of 4,426 trajectories with the split between \textit{out of city center} and \textit{into city center} being 206 and 4,230 respectively.

\subsection{Potential critical \glspl{pvi}}

% Gesamtmenge bei denen temporal overlap. xx (maybe 3.000 ? ) 
% Gesamtmenge bei denen temporal and spatial overlap : (PET < 5) ~1.100 Interactions

From the reconstructed trajectories, 20.978 \glspl{pvi} with a temporal relation result, serve as the baseline catalog. The mean of the absolute \gls{pet} of these \gls{pvi} scenarios was 20.4($\pm$ 25.9)s, the median value is at 14.1s. The distribution in the constellations is 36.6\% for \textit{pedestrian first} and 63.4\% for \textit{vehicle first}. In  54.1\% the vehicle crossed on the \textit{near-side lane} and 45.9\% from the \textit{far-side lane}. 

\begin{figure}[t!]
  \centering
  \includegraphics[width=\columnwidth]{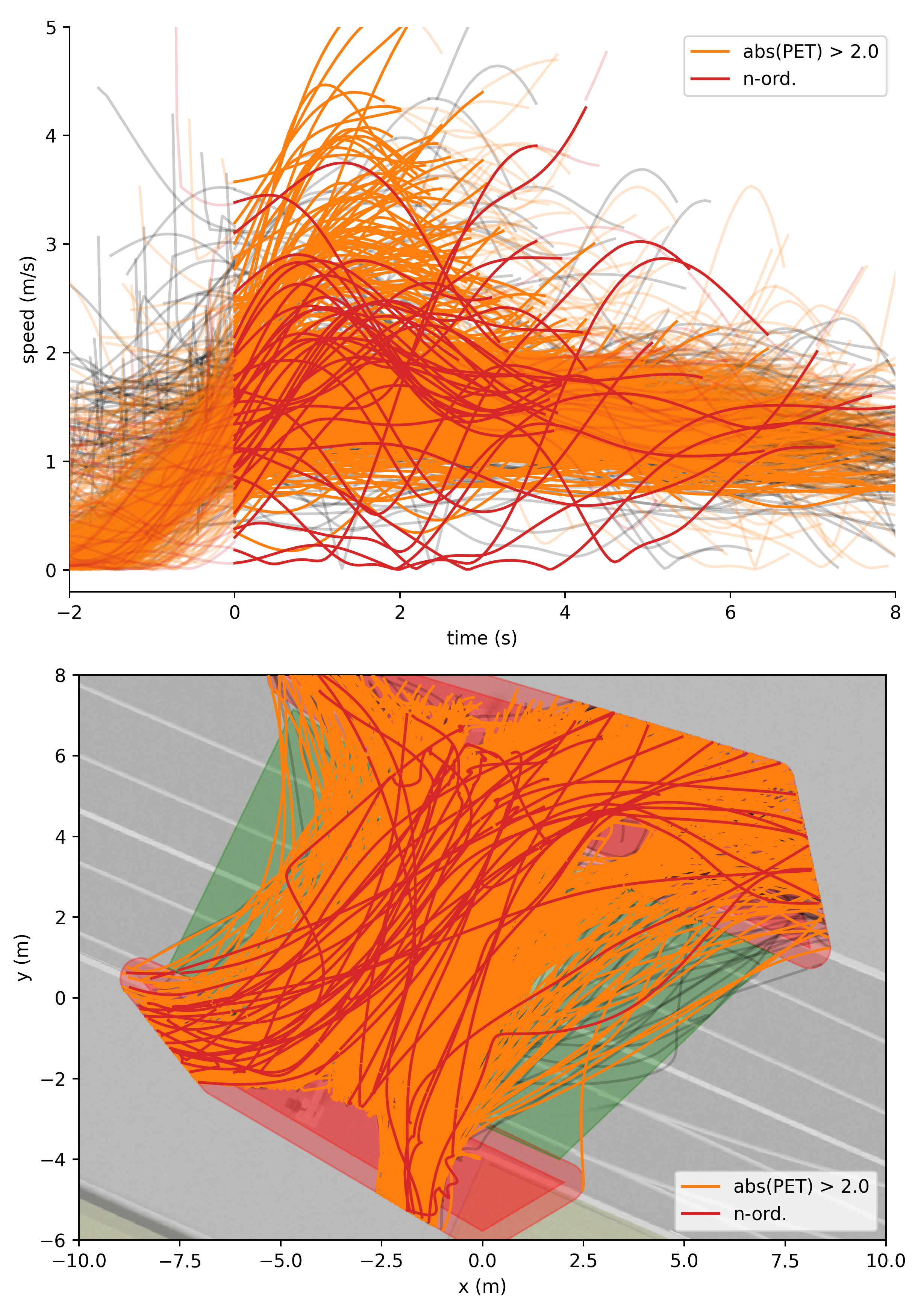}
  \caption{Overview of the extracted scenario catalog of \gls{pcpvi}. The first subfigure shows the pedestrians' speed profiles over time. They are aligned to the time in which they exit the approach zone. The second subfigure shows the reconstructed paths of the pedestrians on the road network. \glspl{pvi} in which the absolute \gls{pet} $\geq$ 2s is black, \gls{pvi} with a non-ordinary motion profile are highlighted red, where the rest is colored orange.}
  \label{fig:reco_speed_profiles}
\end{figure}

The catalog was narrowed down for the further consideration, to \glspl{pvi} in which the \gls{pet} is between -4 and 4s, which we further denote as \textit{\gls{pcpvi}}. This subset consists of a total number of 669 \glspl{pvi}, which make 6.42\% of the analyzed pedestrian trajectories. The distribution between \textit{pedestrian first} and \textit{vehicle first } is 31.6\% and 68.4\%, with a mean \gls{pet} of 2.56~($\pm$ 0.92)s, and -2.31~($\pm$ 1.04)s, respectively. The interaction results in 53.7\% scenarios in which the vehicle crosses on the \textit{near-side lane} and 46.3\% crossing from the \textit{far-side lane}. An outline of the reconstructed \gls{pcpvi} catalog is given in Fig.~\ref{fig:reco_speed_profiles}.

% A qualitative evaluation revealed that this sample still included several scenarios which did not show interesting \glspl{pvi} with motion adaption. 
% Therefore, the proposed \textit{motion adaptation metric} was calculated and grouped into scenarios with $\operatorname{std}(\mathbf{e})$ smaller and larger than 0.03954m/s, which represents the 95\% percentile of all $\operatorname{std}(\mathbf{e})$, with a mean of 0.011 ($\pm$ 0.0187)m/s. 
% In 5\% of the scenarios with absolute \gls{pet} between 2 and 4s this threshold was reached. For the \textit{\gls{pcpvi}} odds were significant higher for reaching the threshold when \gls{pet} was between -2 and 0s. 

\begin{figure}[t!]
  \centering
  \includegraphics[width=\columnwidth]{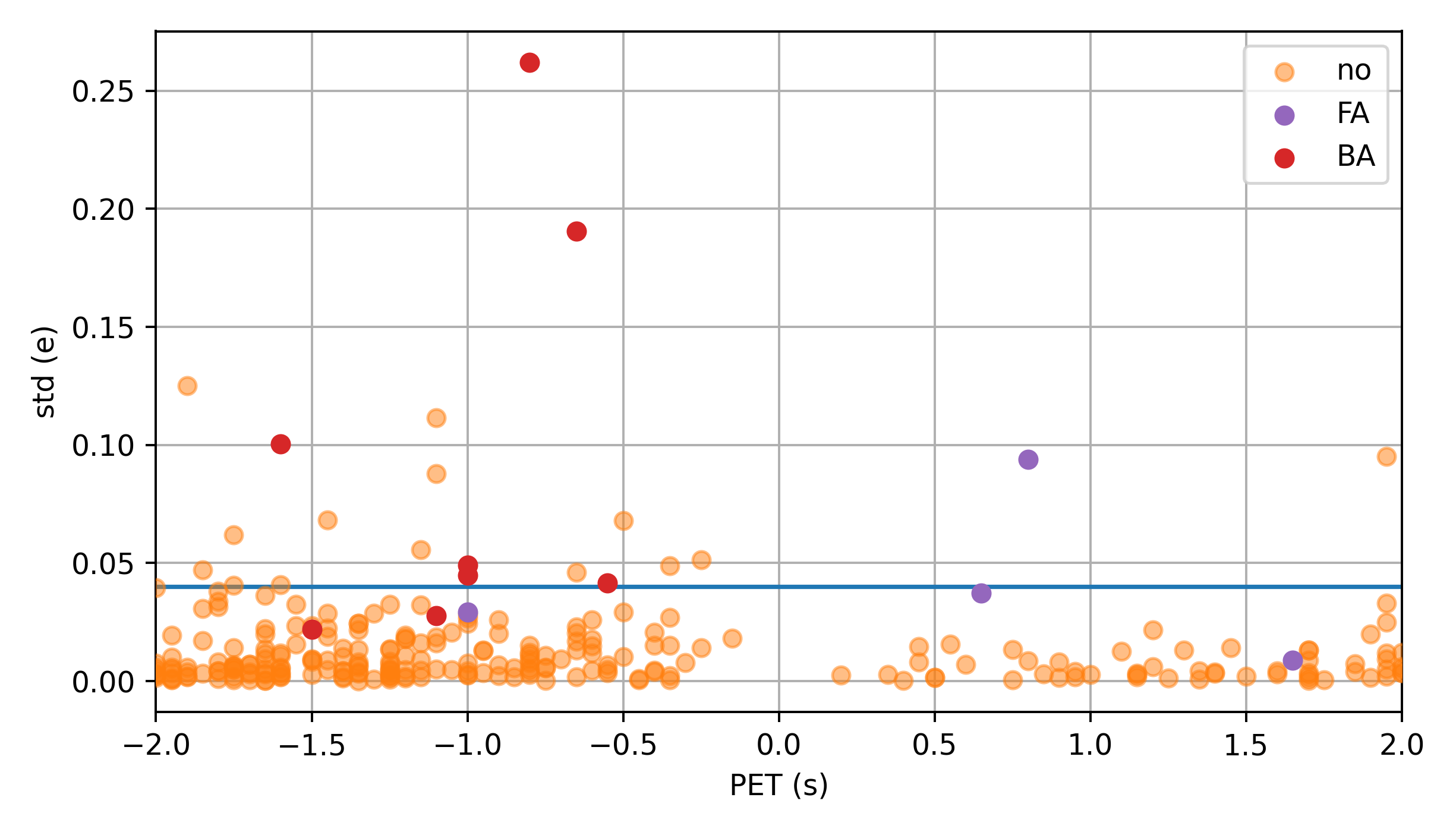}
  \caption{The calculated $\operatorname{std}(\mathbf{e})$ over \gls{pet}, for the considered \gls{pcpvi}, with an absolute \gls{pet} $\leq$ 2s. The colors indicate qualitative analysis of whether the pedestrian showed an motion adaption, identified as backward avoidance (BA) and forward avoidance (FA). }
  \label{fig:pet_vs_std_e}
\end{figure}

The refinement of the catalog with respect to the proposed \textit{motion adaption metric} into ordinary and non-ordinary speed profiles, was carried out by grouping them according to the threshold 0.04m/s, representing approximately the 95th percentile of all $\operatorname{std}(\mathbf{e})$. 
For the \textit{\gls{pcpvi}} odds were significant higher for reaching the threshold when \gls{pet} was between -2 and 0s as summarized in Table~\ref{tab:or_pet_std_e}. A qualitative evaluation, considering all \glspl{pvi} with a absolute \gls{pet} value below 2s showed a false-positive rate of 0.057. The highlighted cases are shown in Fig.~\ref{fig:pet_vs_std_e}

% A qualitative evaluation revealed that this sample still included several scenarios which did not show interesting \glspl{pvi} with motion adaption. 

%we calculated the \gls{or} and the corresponding 95\% confidence interval for each group. We refer to the frequency of cases where $\operatorname{std}(\mathbf{e})$ is below or above the value of 0.04m/s. 

%The \gls{pcpvi} form the basis for further analysis. Based on the \gls{pet} the catalog was further divided into subgroups ([-4 \dots -2], [-2 \dots 0], [0 \dots 2], [2 \dots 4]). Pedestrian trajectories are used as the population here, yielding a ratio of 10544 to 545, and the results are summarized in Table~\ref{tab:or_pet_std_e}. We could not encounter a significant difference for a motion adaption within the overall catalog of \gls{pcpvi},  (OR : 1.23, with 95\% CI : 0.92 -- 1.74). Nevertheless,  the odds for a higher motion adaption for \glspl{pvi} in which the \gls{pet} is in the range of [-2 \dots 0] is significantly higher. For the other groups this is not the case. The properties, by means of \gls{pet}, motion adaption, speed profiles and paths of the refined \gls{pcpvi} catalog are shown in Fig.~\ref{fig:pet_vs_error}.  

\begin{table}[]
    \centering
    \begin{tabular}{l | c | c  |  c | c | c}
    \toprule
        PET Gr.  & Share(\%) & \# ord. & \# n-ord.  & \textbf{OR} & \textbf{95\%- CI} \\ \cmidrule{1-6} 
            -4 \dots -2 & 2.59  & 274  & 13   &  0.92         & 0.52 - 1.61 \\ %\cmidrule{0-0}
            -2 \dots\phantom{-}0  & 1.80   & 181  & 19   & \textbf{2.07} & \textbf{1.28 - 3.34} \\ %\cmidrule{1-1}
            \phantom{-}0 \dots\phantom{-}2   & 0.53   & \phantom{0}57   & \phantom{0}2    &  0.68         & 0.17 - 2.78 \\ %\cmidrule{1-1}
            \phantom{-}2 \dots\phantom{-}4   & 1.50   & 157  & \phantom{0}9    &  1.11         & 0.56 - 2.19 \\ \cmidrule{1-6}
            \gls{pcpvi}  & 6.42 & 669  & 43   &  1.26         & 0.92 - 1.74 \\ 
    \bottomrule
    \end{tabular}
    \caption{\Acrfull{or}, with their 95\% confidence interval (95\% - CI) of pedestrian motion adaptions (ordinary (or.) and non-ordinary (n-or.)) for different \gls{pet} groups (\gls{pet} Gr.) within the \gls{pcpvi}. The share in \% refers to the overall amount of pedestrian trajectories.}
    \label{tab:or_pet_std_e}
\end{table}

% As outlined in Fig.~\ref{fig:pet_vs_error}, many \glspl{pvi} happen with \textit{cyclists} in \textit{pedestrian first} constellations (95 \textit{cyclists} vs 130 \textit{cars}). Considering only this \textit{constellation} in the subcatalog, the mean value for the \textit{cyclist} is at 2.245 ($\pm$ 1.06) s, whereas the mean for \textit{cars} lies at 2.79 ($\pm$ 0.718)s. The Mann-Whitney U test showed a significance (p $<$ 0.05) for those two groups.  
% The boxplot of the distribution is shown in Fig.~\ref{fig:}.

 % \begin{table}[]
 %     \centering
 %     \begin{tabular}{c | c || c | c | c | c | c | c}
 %         		 &       & 1 & 2 & 3 & 4 & 5 & 6  \\ \hline
 %                 & Total & -4..-2 & -2..0 & 0..2 & 2..4 & -2..2 & -4..4 \\ \hline \hline
 %    STD $>$ 0.04 & 545   & 13  & 19  & 2 & 9   & 21  & 43 \\
 %    STD $<$ 0.04 & 10544 & 274 & 181 & 57 & 157 & 238 & 669 \\ \hline
 %    %classified as critical based on expert judgement							
 %               OR & & 0.92 & 2.07 & 0.68 & 1.11 & 1.74 & 1.26 \\
 %    95\% CI lower & & 0.52 & 1.28 & 0.17 & 0.56 & 0.87 & 0.92 \\
 %    95\% CI upper & & 1.61 & 3.34 & 2.78 & 2.19 & 3.45 & 1.74 \\ \hline 
 %     \end{tabular}
 %     \caption{odds ratio}
 %     \label{tab:my_label}
 % \end{table}

\subsection{Critical \glspl{pvi}}

\begin{figure}[!ht]
  \centering
  \includegraphics[width=1.0\columnwidth]{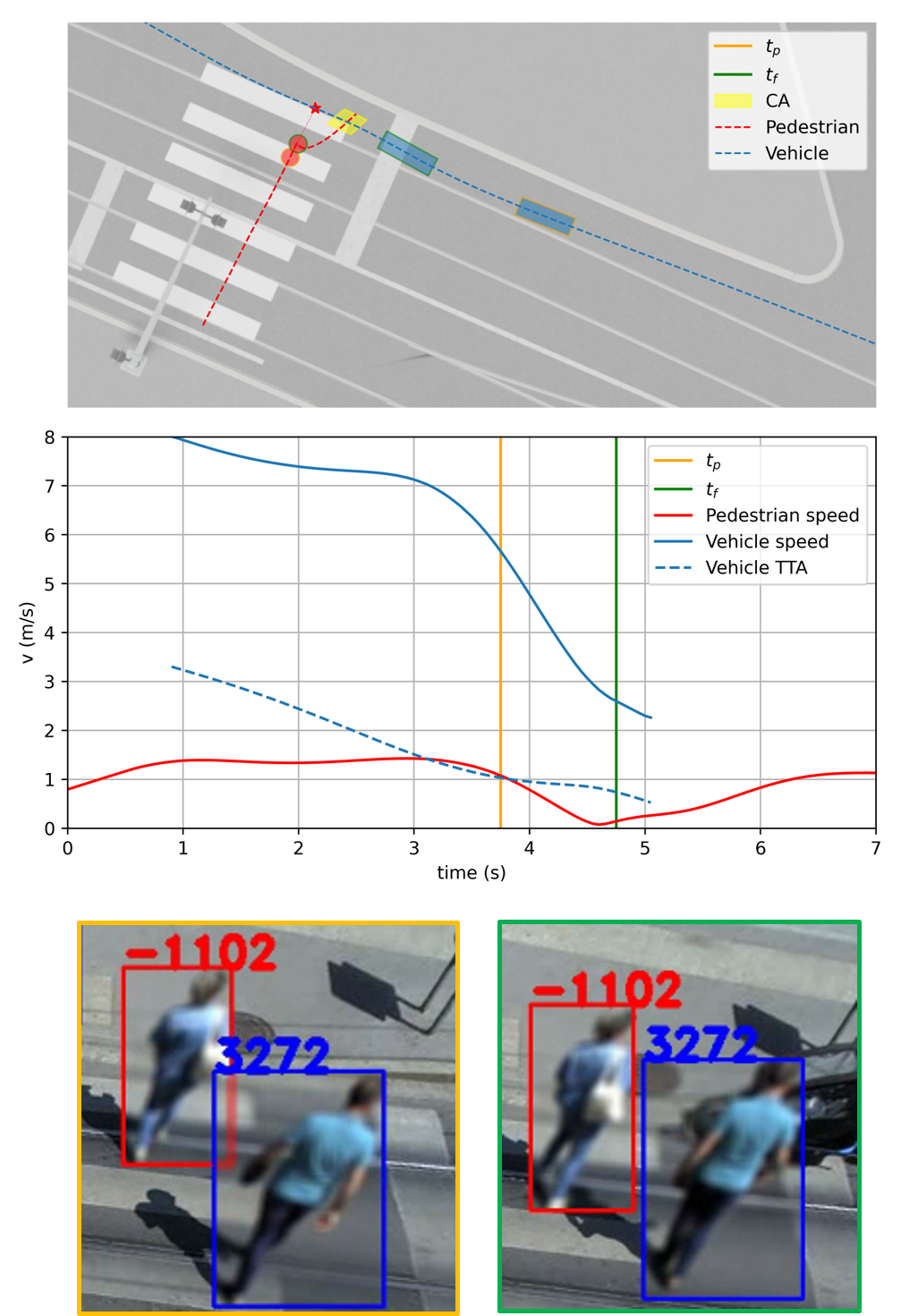}
  \caption{An identified critical \gls{pvi}, where the motion adaption occurs at a relatively late stage. The calculated \gls{pet} is at a value of -0.55s where the $\operatorname{std{(\mathbf{e})}}$ is 0.0416. The visualization shows the position of the vehicle and the pedestrian at time $t_p$, $t_f$ together with the respective frames in the video. In addition both the \Acrfull{ca} and the predicted $\text{CA}^\prime$ are shown. }
  \label{fig:traj_critical_pvi}
\end{figure}

The identified group of critical scenarios, contains 21 cases, which are publicly available at$^3$. Fig.~\ref{fig:traj_critical_pvi} shows the trajectories, speed profiles and the timeline of one identified critical \gls{pvi}, which had a \gls{tta} of 1.05s at the time $t_p$ the pedestrian perceived the vehicle.

%wir eine xx Fälle wurde weiters die \gls{tta} als metrik berechnet, um zu identifizerien ob sich der  Fußgänge

%However, these cases arise mainly for \gls{pvi} of type \textit{vehicle first} in the \gls{ca}. The distribution is shown in Fig.~\ref{fig:pet_vs_error}. 

% From the chosen constraints there are xx scenarios which can be considered as critical and for which the considered constellation works (long enough on the road, going from zone 2 to 1 and vehicle on the far-side). For these scenarios was further examined for their \gls{tta}. For xx of xx cases no \gls{tta} was found, the remaining critical \glspl{pvi} are summarized in \tablename~\ref{tab:critical_pvi}. Figure~\ref{fig:traj_critical_pvi} shows interaction of the most critical scenario catalog.

\section{DISCUSSION}
\label{sec:discussion}

We analyzed long-term traffic recordings and proposed a schema to find a subset of critical \glspl{pvi} in the shown approach here. In a first step we intend to discuss the limitations of the data collections and also of the preprocessing step and furthermore set the proposed schema into the state of the art context.

\subsection{Data collection and preprocessing}
%They reach from benchmarks for tracking algorithms, \emph{e.g.}~\cite{Chavdarova2017}, to first-person vehicle centered data sets incorporating underlying road network \cite{Houston2020, Caesar2020} for \gls{adas} and \gls{ad} development.

Many different pedestrian data sets recorded for specific applications have been published in recent years. These data sets \cite{Robicquet2016, Zhan2019, Bock2020} feature \glspl{pvi} recorded by drones. The camera placement and collected data presented in this study is based on a stationary camera placed closer to the pedestrians and thus allowing the reconstruction of further details, such as intrinsic influencing factors and \gls{pvi} specific attributes such as the perception of the vehicle $t_p$. 
Nevertheless, there are some limitations here due to the setup. The reconstruction accuracy depends on the image area, and the camera placement. Since the observation system was oriented in such manner that the crosswalk is in the center of the recorded frames, the projection can be assumed to be of high accuracy, as investigated in \cite{Schachner2023}. 

Automatic scene reconstruction has limitations, as described in \cite{Kirillova2022}. For this reason we thus utilized the defined \gls{roi} to also identify inconsistencies in reconstructed tracks, i.e. due to hand over issues from one view to another. Trajectories which could not be associated across both camera views have been excluded. The \gls{roi} itself as well as the pedestrian zones have been empirically defined in alignment with the road layout, in a manner similar to the approach as shown in \cite{Gorrini2018}.

\subsection{Criticality assessment}

The classification of the extracted \glspl{pvi} catalog is generally difficult, since the interactive pedestrian behavior depends on various influencing factors as summarized in \cite{Rasouli2020}. 
With regard to criticality assessment, a combination of different metrics have shown a benefit in the past, as for instance shown in  \cite{Cafiso2011, Peesapati2018}.

\subsubsection{Space-sharing conflict}

In comparison to the \gls{ssc} definition in \cite{Markkula2020}, the \gls{ssc} was interpreted generously in our study. The authors describe the term \gls{ssc} as a traffic situation in which the vehicle and the pedestrian would occupy the same place at the same time. Theoretically, this characterization always has a prediction of some kind in its definition. In our study, \gls{pet} was used as a metric to determine this temporal and spatial proximity of both the pedestrians and the vehicle. The simple temporal proximity, however, yields a large baseline catalog. 
For the identification of a \gls{pcpvi} subset the threshold of $<$4s was used. The authors in \cite{Johnsson2018} summarize different studies which assess the capability of the\gls{pet} for identifying \gls{ssc}. The common threshold lies in the range between $<$1-3s also depending on the application. Nevertheless, it is also admitted that there might be critical scenarios in which \gls{pet} fails, for example, when the vehicle induced an emergency brake and fully stops in front of the pedestrian. 

Further, multiple \glspl{pvi} were determined for each pedestrian using our approach. Nevertheless, we considered the \gls{pvi} with the lowest absolute \gls{pet} as the one with the highest impact on the pedestrian behavior, which might not always be the case. This was observed in examples where the pedestrian was waiting in the approach zone, crossing behind a vehicle on the near-side lane or slowing down due to a vehicle on the far-side lane. 
The given examples underpin that \gls{pet} yields also to a broad number of constellations which do not entail further criticality.

\subsubsection{Motion adaption}

The adaptation of the speed, for example by braking, is one of the most common driving actions to avoid a collision. Therefore, metrics have been postulated in the past that consider the deceleration of the vehicle as an indication of a critical situation \cite{Hyden1987}. To the author's knowledge, the described pedestrian deceleration by \cite{Nie2021, Li2021} has so far not been considered as a metric to assess the criticality of a scenario. As shown in Fig.~\ref{fig:pet_vs_std_e}, the proposed motion adaption metric can be attributed to \glspl{pvi}, which have been quantitatively identified as critical. The deviation to an approximated second order polynomial would appear to be a good indicator and one that could also compliment other indicators. 
In addition to this, the deviation between the predicted CA' and the actual CA, as shown in Fig.~\ref{fig:traj_critical_pvi}, might prove to be a useful additional indicator for evaluating the evasive behavior of pedestrians. 

Nevertheless, the authors in \cite{Nie2021, Li2021}, also describe forward avoiding behavior, i.e. the acceleration in the immediate vicinity of the vehicle. The metric described here works less well for the \glspl{pvi} of these \textit{pedestrian first} constellations, which might be due to the fact that accelerations only, could be better approximated through a second order polynomial. Furthermore, the acceleration might be more difficult to distinguish from other factors than the proximity of the vehicle, such as rushing to catch a bus or a running phase, because of jaywalking \cite{Tian2013}.

% In this study, the motion adaptation was drawn from the findings of \cite{Li2021, Nie2021} which analyzed pedestrian behavior in critical situation using volunteer tests. 
% They conclude that upon the delayed perception a reaction takes place, which can be seen as some kind of escaping approach, denoted by a changing speed profile. The studies conclude that pedestrians typically show certain reactions, such as stepping back or forward, if they perceive the vehicle.
% In our case, we rely on the described backward avoidance outlined in Fig.~\ref{fig:curve_fitting} as a criterion for identifying the motion adaption and thus the interactive behavior. 

\subsection{Scenario catalogs}

In the current data set of 110h, around 700 cases remained, a result which would appear to be reasonable due to the fact that severe conflicts are relatively rare. In the recorded data of \cite{Peesapati2013}, for instance, a 10 hour period yielded only 74 conflicts with \glspl{pet} under 3s. 
Longer recordings would also provide an opportunity to further investigate influencing factors such as age, gender or group effects \cite{Rasouli2020}. 
Furthermore, it could be applied to other data sets such as \cite{Bock2020, Zhan2019}, which would further verify the capabilities of the methodology.

From identified \glspl{pvi} scenarios, important quantities to utilize  behavioral pedestrian models can be drawn \cite{Yang2020}. Besides, \gls{tta} and desired walking speeds, the time at which the pedestrian perceives the vehicle represents an important characteristic \cite{Nie2021} and was presented for the scenario in Fig.~\ref{fig:traj_critical_pvi}. An incorporation into the development process of \gls{adas} and \gls{ad} functions \cite{Schachner2020, Neurohr2020} would therefore be a benefit.

In addition to this the \gls{pcpvi} catalog offers a useful data source for further investigations. The \gls{pvi} outlined in Fig.~\ref{fig:traj_critical_pvi} the conflict vehicle was a bicycle, which could also be identified for other critical \gls{pvi}. It would appear that \glspl{pvi} with cyclists occur at a smaller \gls{pet} compared to cars, which again appears to be in line with \cite{Pawar2015}. They postulated an influence of different motorized vehicle types on gap acceptance, which increases with vehicle size, but which still needs further investigation. 

In the course of our manual inspections, we identified many traffic violations. The speed profiles in Fig.~\ref{fig:reco_speed_profiles} also indicate that many people are rushing, which might be due to factors such as willingness to jaywalk as also identified in \cite{Tian2013}, the dash to catch a bus about to depart, or also -the response to oncoming traffic \cite{Nie2021}.

\section{CONCLUSION}
\label{sec:conclusion}

With the demonstrated approach, 21 critical \gls{pvi} scenarios have been identified by using the camera-based observation system at the selected crossing. The introduced method to automatically filter for critical \gls{pvi} combining the \gls{pet} as \gls{ssc} metric with pedestrian motion adaptation is a promising approach for handling large volumes of recordings from traffic observations. The described data set of critical \gls{pvi} was published$^3$ to enable researchers to calibrate and validate behavioral pedestrian models in the scenario-based development of safety systems. 

\bibliography{bibliography} 

\bibliographystyle{ieeetr}

\end{document}